\documentclass[12pt]{article}
\usepackage{arxiv}
\usepackage{algorithm,algorithmic}
\usepackage{amsmath,amssymb,amsfonts}
\usepackage{amsmath,amsfonts,bm}

\def\eqref#1{equation~\ref{#1}}
\def\1{\bm{1}}

\def\eps{{\epsilon}}

\def\rvepsilon{{\mathbf{\epsilon}}}

\def\rva{{\mathbf{a}}}
\def\rvb{{\mathbf{b}}}
\def\rvc{{\mathbf{c}}}
\def\rvd{{\mathbf{d}}}
\def\rve{{\mathbf{e}}}

\def\rvh{{\mathbf{h}}}
\def\rvu{{\mathbf{i}}}

\def\rvm{{\mathbf{m}}}
\def\rvn{{\mathbf{n}}}
\def\rvo{{\mathbf{o}}}
\def\rvp{{\mathbf{p}}}

\def\rvr{{\mathbf{r}}}

\def\rvu{{\mathbf{u}}}
\def\rvv{{\mathbf{v}}}

\def\rvx{{\mathbf{x}}}
\def\rvy{{\mathbf{y}}}
\def\rvz{{\mathbf{z}}}

\def\vmu{{\bm{\mu}}}
\def\vtheta{{\bm{\theta}}}

\def\mI{{\bm{I}}}

\def\mU{{\bm{U}}}

\def\mW{{\bm{W}}}

\def\mSigma{{\bm{\Sigma}}}

\DeclareMathAlphabet{\mathsfit}{\encodingdefault}{\sfdefault}{m}{sl}
\SetMathAlphabet{\mathsfit}{bold}{\encodingdefault}{\sfdefault}{bx}{n}

\def\sR{{\mathbb{R}}}

\DeclareMathOperator*{\argmin}{arg\,min}

\usepackage{todonotes}
\usepackage{float}
\usepackage{multicol,multirow}
\usepackage{secdot}
\usepackage{amsmath,array,ragged2e}
\newcolumntype{C}{>{\Centering\arraybackslash}m{0.14\linewidth}}
\usepackage{booktabs}
\usepackage{adjustbox}
\usepackage{subcaption}
\usepackage{graphicx}
\usepackage{hyperref}
\graphicspath{{./pics/}}
\usepackage{lineno}
\usepackage{tabularx}

\def\eqref#1{(\ref{#1})}

\title{Memory-efficient particle filter recurrent neural network for object localization}
\author{Roman Korkin \\ 
Novosibirsk Technology Center \\
Schlumberger, Russia\\
\url{korkin.rv@phystech.edu}
\And 
Ivan Oseledets \\
AIRI, Moscow, Russia \\
Skoltech, Moscow, Russia \\
\url{i.oseledets@skoltech.ru}
\And Aleksandr Katrutsa\thanks{Corresponding author}\\
Skoltech, Moscow, Russia \\
AIRI, Moscow, Russia \\
\url{aleksandr.katrutsa@phystech.edu}
}

\begin{document}

\maketitle

\begin{abstract}
    This study proposes a novel memory-efficient recurrent neural network (RNN) architecture specified to solve the object localization problem.
This problem is to recover the object states along with its movement in a noisy environment.
We take the idea of the classical particle filter and combine it with GRU RNN architecture.
The key feature of the resulting memory-efficient particle filter RNN model (mePFRNN) is that it requires the same number of parameters to process environments of different sizes. 
Thus, the proposed mePFRNN architecture consumes less memory to store parameters compared to the previously proposed PFRNN model.
To demonstrate the performance of our model, we test it on symmetric and noisy environments that are incredibly challenging for filtering algorithms.
In our experiments, the mePFRNN model provides more precise localization than the considered competitors and requires fewer trained parameters.
    % This problem has specific features that allow to modify of the general purpose GRU PFRNN, reduce the number of parameters, and preserve or even improve the localization precision.
    % The proposed filtering algorithm based on our cell of the recurrent neural network is further called model-based particle filter recurrent neural network (MB-PFRNN). 
    % We test the proposed GRU cell in symmetric and noisy environments that are especially challenging for filtering algorithms.
\end{abstract}

\section{Introduction}
We consider the object localization problem and propose a novel GRU-like architecture to solve it.
Typically standard GRU-like models~\cite{shi2018machine} are used to process sequential data, e.g. to predict the next item in a sequence, and classify or generate texts, audio, and video data.
The object localization problem differs from the aforementioned problems since auxiliary data about the environment and particular measurements are available.
Therefore, this additional knowledge should be incorporated into the GRU architecture properly.
Such a modification can be based on the existing approaches to solve the object localization problem, which are discussed further.

One of the classical non-parametric methods to solve the object localization problem is particle filter~\cite{gustafsson2010particle}, which estimates the filtered object state from the states of auxiliary artificial objects that are called particles.
A modification of GRU and LSTM recurrent neural networks with particle filter ingredients is presented in~\cite{ma2020particle}, where a particle filter recurrent neural network (PFRNN) is proposed.
The core element of PFRNN is the modified cell (GRU or LSTM) equipped with analogs of particles and the corresponding weights of particles to estimate the filtered state.
However, PFRNN improves the performance of the general sequential data processing and does not consider specific features of the object localization problem.
Therefore, we propose the novel \emph{memory-efficient PFRNN (mePFRNN)} that combines the model assumptions used in the classical filtering methods (e.g. Kalman filter and particle filter) and parametrization from the GRU architecture.
Such a combination provides more accurate state estimation and improves robustness in noisy and symmetric environments.
Also, the Soft resampling procedure is used to avoid the degeneracy issue and improve the stability of the filtered states.

The main contributions of our study are the following.
\begin{enumerate}
    \item We propose a modification of the PFRNN architecture specified for the object localization problem. The proposed mePFRNN model does not exploit environment embeddings and extracts this data implicitly in the training stage. %explicitly takes into account the measurements and beacons' locations. 
    \item We perform an extensive experimental comparison of the proposed GRU-like architecture with the existing recurrent neural networks and other non-parametric methods like the particle filter.
    \item The proposed mePFRNN model requires the same number of parameters for environments of different sizes.
\end{enumerate}

\paragraph{Related works.}
The object localization problem appears in a lot of applications like driving autonomous vehicles~\cite{woo2018localization}, navigation~\cite{barczyk2012invariant,lim2012design}, image processing~\cite{costagli2007image}, finance~\cite{racicot2010forecasting} and fatigue predictions~\cite{yang2017application}.
Therefore, there are a lot of different approaches to solving it.
We can split them into two classes: non-parametric and parametric.
The first class consists of classical methods that do not require a training  stage and perform filtering of the object states on the fly.
Examples of such methods are Kalman filter~\cite{auger2013industrial,grewal2010applications},  and its modifications like extended~\cite{julier1997new}, unscented~\cite{julier2004unscented}, invariant extended~\cite{bonnable2009invariant} and ensembled~\cite{houtekamer1998data} Kalman filters.
Also, methods related to the particle filter, e.g. multiparticle Kalman filter~\cite{korkin2023multiparticle}, particle filters combined with genetic algorithms~\cite{moghaddasi2020hybrid}, and particle swarm technique~\cite{particle_swarm}, box particle filter~\cite{gning2013particle} and others are non-parametric filtering methods.
The second class consists of parametric methods such that  a pre-training stage is necessary before starting filtering.
Such methods are typically based on neural networks that are trained on the collected historical data and then tested on the new data from real-world simulations.
Although the pre-training stage may require a lot of time, one can expect that the inference stage, in which filtering is performed, is sufficiently fast due to modern hardware acceleration.
Moreover, since the neural network models can efficiently treat sequential data~\cite{wang2020deep,lecun2015deep}, the parametric methods can provide more accurate filtering results compared to non-parametric methods.

Although the Transformer model~\cite{vaswani2017attention} demonstrates superior performance over the considered GRU RNN in sequence processing tasks, it consumes a lot of memory to store parameters, requires special techniques for training~\cite{gusak2022survey} and may not fit in the on-device memory limits.
The memory-efficient Transformer models~\cite{wang2020linformer,kitaev2020reformer,jaszczur2021sparse} may be a remedy for the observed issue and will be investigated in future work. 
\section{Problem statement}

Consider the trajectory of object states encoded as a sequence of $d$-dimensional vectors $\rvx_i \in \sR^d$, where $i$ is an index of the time moment $t_i$.
For example, if the object's state consists of 2D coordinates and 2D velocity, then state dimension $d=4$.
The states are changed according to the motion equation, which combines the physical law and the control system of the object.
Formally we can write the motion equation as follows
\begin{equation}
    \rvx_i = f(\rvx_{i - 1}, \rvu_{i}, \boldsymbol{\eta}_i),
    \label{eq::motion_func}
\end{equation}
where $\rvu_i$ is a vector of control at the time moment $t_i$, for example, external forces, and $\boldsymbol{\eta}_i$ is a vector of noise corresponding to the object motion at the time moment $t_i$.
Since the object moves with some noise, we should use additional measurements to estimate states more precisely.
Typically there are several beacons in the environment, which are used by objects to measure some quantities that can improve their state estimate.
For example, distance to the $k$-nearest beacons can improve the estimate of the object's location.
Formally, denote by $\rvy_i \in \sR^k$ a vector of measurements at time moment $t_i$ that is related with state estimate through the measurement function $g: \mathbb{R}^d \rightarrow \mathbb{R}^k$:
\begin{equation}
\rvy_i = g(\rvx_i, \boldsymbol{\zeta}_i),
\label{eq::measure_func}
\end{equation}
where $\boldsymbol{\zeta}_i$ is the additional noise of measurement. 

Object localization problem is the problem of estimating object trajectory from the given motion and measurement functions that represent the physical law of the environment and beacons' configuration, respectively.
In this study, we introduce the parametric model $h_{\vtheta}: \sR^d \times \sR^k \times \sR^n \to \sR^d$ that depends on the unknown parameters $\vtheta \in \sR^n$ and performs filtering of the inexact state estimate $\rvx_i$ based on the additional measurements $\rvy_i$.
Assume we have training trajectory of the ground-truth states $\{\rvx_i^*\}_{i=1}^N$.
Then we can state the optimization problem to fit our parametric model to the training data $\{\rvx_i^*\}_{i=1}^N$ and evaluate the generalization ability of the resulting model.
In particular, the standard loss function in such a problem is the mean square error loss function 
\begin{equation}
MSE = \frac{1}{N}\sum_{i=1}^{N}{\|h_{\vtheta}(\rvx_i, \rvy_i)-\rvx^{*}_i\|_2^2}
\label{eq::mse_def}
\end{equation}
such that the motion function $f$ and the measurement functions $g$ give the state estimate $\rvx_i$ and measurement vector $\rvy_i$, respectively.

We further focus on the plane motion setup, where the state vector consists of 2D coordinates $\rvc \in \mathbb{R}^2$ and a heading $\alpha \in [0, 2\pi]$, which defines the direction of movement, i.e.  $\rvx = [\rvc, \alpha] \in \mathbb{R}^3$.
Therefore, we follow~\cite{ma2020particle} in slightly adjusting the MSE objective function~\eqref{eq::mse_def} to treat coordinates and angles separately and compose the weighted MSE loss function:
\begin{equation}
    wMSE = \underbrace{\frac{1}{N} \sum_{i=1}^{N}\|\rvc_i-\rvc^{*}_i\|_2^2}_{=\mathrm{MSE}_c} + \frac{\beta}{N} \sum_{i=1}^{N} (\alpha_i - \alpha_i^*)^2, \quad \text{where} \quad [\rvc_i, \alpha_i] = h_{\vtheta}(\rvx_i, \rvy_i), \quad \rvx_i^* = [\rvc_i^*, \alpha_i^*],
    \label{eq::wmse_def}
\end{equation}
where $\beta > 0$ is a given weight. 
However, the $wMSE$ loss function treats angles $2\pi-\eps$ and $\eps$ as essentially different while they are physically close.
Thus, we propose a novel modification of the mean squared loss function~\eqref{eq::mse_def}, that treats headings differently.
In particular, we compare not angles but their sine and cosine in the following way:
\begin{equation}
    L(\vtheta) = \mathrm{MSE}_c + \frac{\beta}{N} \sum_{i=1}^N \left[(\sin\alpha_i-\sin\alpha^*_i)^2+(\cos\alpha-\cos\alpha_i^*)^2\right],
    \label{eq::L_def}
\end{equation}
where we use the same notation as in~\eqref{eq::wmse_def}.
Thus, we have the following optimization problem:
\begin{equation}
    \begin{split}
&\vtheta^* = \argmin_{\vtheta} L(\vtheta),\\
\text{s.t. } & \rvx_i = f(\rvx_{i - 1}, \rvu_{i}, \boldsymbol{\eta}_i)\\
& \rvy_i = g(\rvx_i, \boldsymbol{\zeta}_i).
\end{split}
\label{eq:problem_statement}
\end{equation}
% Initial state is random
Additionally to the MSE-like loss function, we evaluate the resulting model with the Final State Error (FSE) loss function, which reads as
\begin{equation}
\mathrm{FSE} = \| \rvc_N - \rvc_N^* \|_2,
\label{eq::fse_def}
\end{equation}
where $[\rvc_N^*, \alpha_N^*] = \rvx_N^*$, 
$[\rvc_N, \alpha_N] = h_{\vtheta}(\rvx_N, \rvy_N)$ and $t_N$ is a last-time moment in the considered period.
Although the FSE loss function is widely used in previous studies~\cite{ma2020particle,zhu2020towards}, it may overestimate the filter performance due to the uncertainty in the filtering process.
The final coordinates may be filtered very accurately by accident while filtering the previous coordinates may be quite poor.
Thus, we focus on the MSE$_c$ loss function as the main indicator of the filter performance.

The key ingredient of this approach is the selection of the proper parametric model $h_{\vtheta}$.
Following~\cite{ma2020particle} we modify the GRU model such that it solves the object localization problem specifically.
A detailed description of our modification is presented in the next section.
\section{Particle filter}

One of the most efficient non-parametric approaches to solving the localization problem is the particle filter.
This filter considers artificially generated particles with states $\rvp^{(k)}_i \in \mathbb{R}^d$ at the $i$-th time step and the corresponding weights $w^k_i \geq 0, \sum_{k=1}^K w^k_i = 1$ such that the estimate of the object state at the $i$-th time step is computed as follows
\[
\hat{\rvx}_i = \sum_{k=1}^K w^k_i \rvp^{(k)}_i,
\]
where $K$ is the number of particles.
Particles' weights are updated according to the corresponding measurements and state updates based on the Bayes rule and likelihood estimation, see~\cite{chen2003bayesian} for details. 
The important step in the particle filter is resampling, which corrects the updated particle weights and states to improve the accuracy of estimate $\hat{\rvx}_i$.
The resampling step addresses the degeneracy issue, which means a few number of particles have non-zero weights.
This phenomenon indicates the poor representation of the target object state.
The purely stochastic resampling samples particles' indices from the multinomial distribution according to the updated weights and then update particle states, respectively, see~(\ref{eq::stoch_resampling}).
After resampling the resulting particle states are slightly perturbed with random noise to avoid equal particles' states.
\begin{equation}
\begin{split}
    &i_1,\ldots,i_K \sim \mathrm{Multinomial}(w^1_{i+1},\ldots,w^K_{i+1})\\
    &\rvp^1_{i+1},\ldots,\rvp^K_{i+1} \leftarrow  \rvp^{i_1}_{i+1},\ldots,\rvp^{i_K}_{i+1}\\
    &w^k_{i+1}=\frac{1}{K}.
\end{split}
\label{eq::stoch_resampling}
\end{equation}
Since the particle filter processes sequential data through the recurrent updates of the particles and weights, the natural idea is to incorporate a similar approach in the recurrent neural network architecture.
The particle filter recurrent neural network is proposed in~\cite{ma2020particle} and we briefly describe it in the next section to highlight the difference with the proposed mePFRNN. 
\section{Recurrent neural networks inspired by particle filter}

This section presents our RNN cell based on the particle filter idea, explicitly measured data, and beacons' positions.
Since our model is a modification of the PFRNN~\cite{ma2020particle} model, we briefly provide the main ingredients of this model.

\paragraph{PFRNN.}
Denote by $K$ a number of particles that are emulated in the PFRNN model.
Below we consider motion $\rvx_i^{(k)}$ and measurement $\rvy_i^{(k)}$ vectors corresponding to the $k$-th particle at the $i$-th time moment, so $k=1,\ldots,K$ and $i=1,\ldots, N$. 
PFRNN considers the environment as a 2D array and constructs its embedding through the following encoder subnetwork:
\begin{equation}
\mathrm{Conv} \to \mathrm{ReLU} \to \mathrm{Conv} \to \mathrm{ReLU} \to \mathrm{Conv} \to \mathrm{ReLU} \to \mathrm{Flatten} \to \mathrm{Linear} \to \mathrm{ReLU},
    \label{eq::env_embed_pfrnn}
\end{equation}
where $\mathrm{Conv}$ is a convolution layer, $\mathrm{ReLU}$ denotes element-wise ReLU non-linearity, $\mathrm{Linear}$ denotes a linear layer and $\mathrm{Flatten}$ denotes a vectorization operation that reshapes the input tensor to a vector.
The output of this subnetwork is the environment embedding vector~$\rve_{env}$.
At the same time, the embeddings for observations $\rvy_i^{(k)}$ and motions $\rvx_i^{(k)}$ are constructed via two linear layers and followed by ReLU activations and denoted by $\rvn_i^{(k)}$ and $\rvm_i^{(k)}$, respectively.
Note that the dimensions of $\rvn_i^{(k)}$ and $\rvm_i^{(k)}$ are the same.
Then, one transforms the environment embedding $\rve_{env}$ to adjusted embeddings $\hat{\rvn}_i^{(k)}$ and $\hat{\rvm}_i^{(k)}$ via two linear layers and followed ReLU activations.
Now, the dimensions of $\hat{\rvn}_i^{(k)}$, $\hat{\rvm}_i^{(k)}$, $\rvn_i^{(k)}$ and $\rvm_i^{(k)}$ are the same.
Finally, the input to the PFRNN cell described below (see~\eqref{eq::pfrnn}) is a set of vectors $\rvv_i^{(k)}$ composed by concatenation of vectors $\hat{\rvn}_i^{(k)} \odot \rvn_i^{(k)}$ and $\hat{\rvm}_i^{(k)} \odot \rvm_i^{(k)}$, where $\odot$ denotes element-wise product. 
% The procedure for computing $\rvv_i^{(k)}$ is summarized in scheme~(\ref{eq::pfrnn_input}).
% \begin{equation}
%     \begin{array}{cc}
%           & \mathrm{Linear} \to \mathrm{ReLU} \to  \\
%           \rve_{env} & \\
%          & \mathrm{Linear} \to \mathrm{ReLU} \to
%     \end{array}
%     \label{eq::pfrnn_input}
% \end{equation}

The baseline PFRNN cell is presented in both a graphical way (see Figure~\ref{fig:pfrnn}) and an analytical way (see equation~(\ref{eq::pfrnn})) for the reader's convenience.
We note that this cell includes a reparametrization trick and updates not only the hidden states for every particle $\rvh_i^{(k)}$ but also the corresponding weights~$w_i^{(k)}$ that are used in the resampling step.
These weights typically correspond to the probability of the particle being equal to the ground-truth object state.
However, in our experiment such weights are the logarithm of the corresponding probabilities, therefore the normalization step after update has the given form (see the last line in~(\ref{eq::pfrnn}).
After that, we adjust the resampling step to deal with the logarithms of the weights properly, see the paragraph below.

\paragraph{Resampling procedure.}
After the inference stage in the considered RNN cells, one has to make resampling, to mitigate the potential degeneracy problem.
There are different approaches to performing resampling~\cite{li2015resampling,towards}.
The main requirement for the resampling procedure in the parametric model is to be differentiable.
Therefore, the stochastic resampling~(\ref{eq::stoch_resampling}) is not directly fitted to the considered model.
Instead, the Soft Resampling procedure~\cite{ma2020particle} was proposed as a trade-off between the accuracy and the related costs. 
This approach to resampling considers a mixture of the distribution induced by weights and the uniform distribution with probabilities $1/K$.
Therefore, the formula for updating weights and hidden states reads as follows.
\begin{equation}
    \begin{split}
        &i_1,\ldots,i_K \sim \mathrm{Multinomial}(\alpha w^1_{i+1} + (1 - \alpha) / K,\ldots, \alpha w^K_{i+1} + (1 - \alpha) / K)\\
    &\rvh^1_{i+1},\ldots,\rvh^K_{i+1} \leftarrow  \rvh^{i_1}_{i+1},\ldots,\rvh^{i_K}_{i+1}\\
    &w^k_{i+1} \leftarrow \frac{w^{i_k}_{i+1}}{\alpha w^{i_k}_{i+1} + (1-\alpha)/K},
    \end{split}
    \label{eq::soft_resamplimg}
\end{equation}
where $\alpha > 0$ to make the operation differentiable.
Note that similar to the stochastic resampling, the updated hidden states $\rvh^{k}_i$ are slightly perturbed.
Section~\ref{sec::experiment} provides more details on the usage of soft resampling in our experiments.

% First, the indices of resampled particles $i{'}$ are multinomially chosen from the original indices $i$ with the probabilities: $\alpha w_i + (1-\alpha)/N$.

% Then, the updated hidden states $h{'}$ are determined through $h$ as $h_{i^{'}}=h_i$. Further, the updated probabilities $w_{i^{'}}$ are: 
% \[
% w_{j^{'}} \leftarrow \frac{w_{j^{'}}}{\alpha w_{j^{'}} + (1-\alpha) 1/N},
% \]
% where $\alpha > 0$ to make the operation differentiable.
% Section~\ref{sec::experiment} provides more details on the usage of soft resampling in our experiments.

\begin{minipage}{.45\textwidth}
\centering
\includegraphics[width=\textwidth]{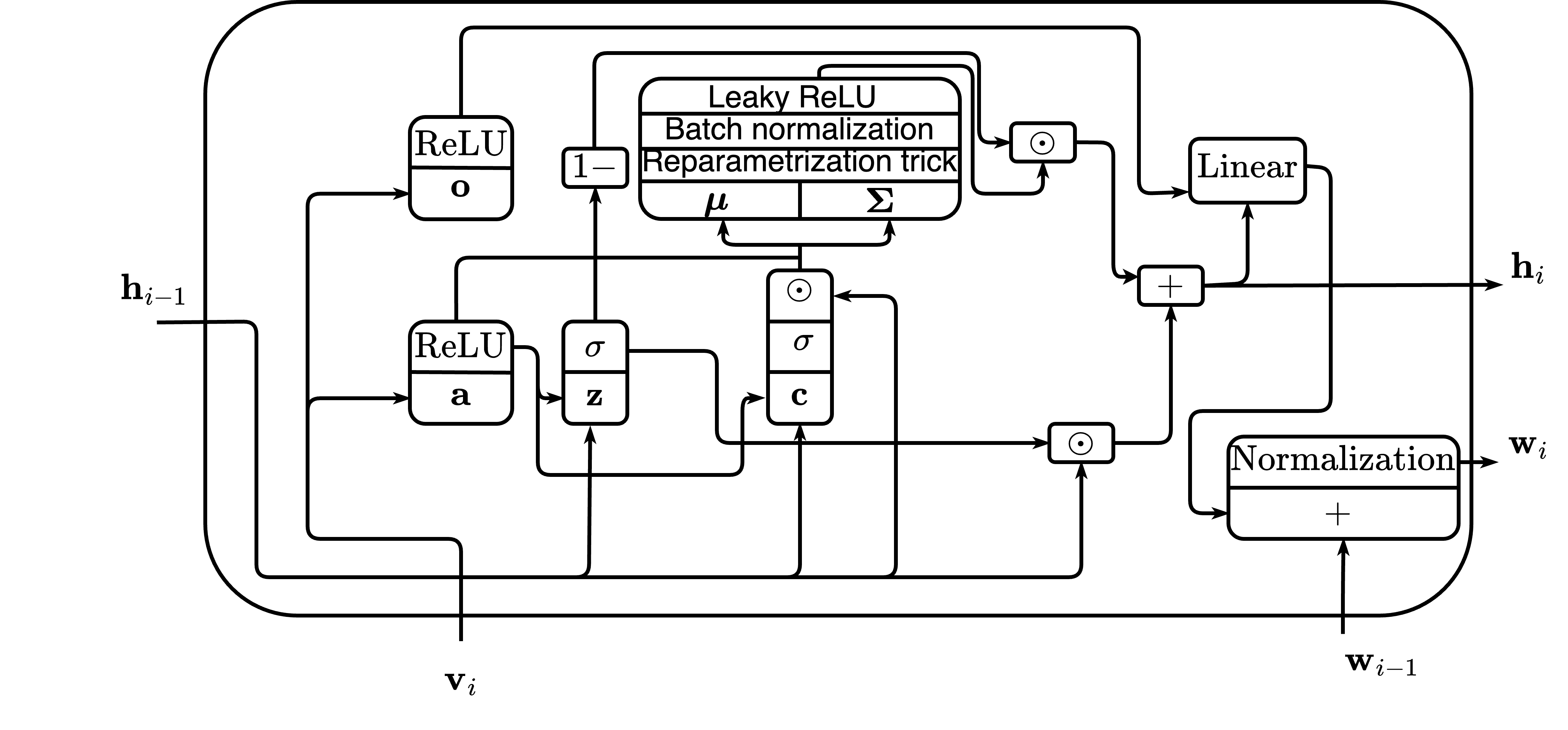}
\captionof{figure}{Baseline PFRNN cell design, where $\mathbf{a},\mathbf{o},\mathbf{z},\mathbf{c}$ denote  the linear layers that are used in computing the corresponding intermediate embeddings. For simplicity, we skip the superscript $k$ that indicates the particle index.
This cell updates both particle hidden states and weights.
Denote elementwise addition and multiplication by $+$ and $\odot$.}
\label{fig:pfrnn}
\end{minipage}
~
\begin{minipage}{.55\textwidth}
\begin{equation}
  \begin{aligned}
  \rva^{(k)}_i =& \mathrm{ReLU}(\mW_a \rvv^{(k)}_i + \rvb_a)\\
  \rvo^{(k)}_i =& \mathrm{ReLU}(\mW_o \rvv^{(k)}_i + \rvb_o) \\
  \rvc_i^{(k)} =& \sigma(\mW_c [\rva^{(k)}_i, \rvh_{i-1}^{(k)} ] + \rvb_c) \odot \rvh_{i-1}^{(k)}\\
  \rvz^{(k)}_i =& \sigma(\mW_z [\rva^{(k)}_i, \rvh_{i-1}^{(k)}] + \rvb_z)\\
  \vmu^{(k)}_i =& \mW_{\vmu} [\rva^{(k)}_i, \rvc_i^{(k)}] + \rvb_{\vmu}& \\
  \mSigma^{(k)}_i =& \mW_{\mSigma} [\rva^{(k)}_i, \rvc_i^{(k)}] + \rvb_{\mSigma}& \\
  \rvepsilon \sim & \; \mathcal{N}(0, \mI)\\
  \rvd^{(k)}_i = & \mathrm{LeakyReLU} (\mathrm{BN}(\vmu^{(k)}_i + \mSigma^{(k)}_i \odot \rvepsilon))\\
  %\rvh_i^{(k)} =& (1 - \rvz^{(k)}_i) \odot \rvh_{i-1}^{(k)} + \rvz^{(k)}_i \odot \rvd^{(k)}_i  \\
  \rvh_i^{(k)} = & (1 - \rvz^{(k)}_i) \odot  \rvd^{(k)}_i + \rvz^{(k)}_i \odot \rvh_{i-1}^{(k)}  \\
  \rvp_i^{(k)} =& \mW_w [\rvo_i, \rvh_i^{(k)}] + b_w \\ 
  w_i^{(k)} = & \rvp_i^{(k)} + w_{i-1}^{(k)} - \\
  & - LogSumExp( \rvp_i^{(k)} + w_{i-1}^{(k)} ) 
  \end{aligned} 
  \label{eq::pfrnn}
  \end{equation}
  \end{minipage}

\paragraph{mePFRNN.}
Since PFRNN encodes the environment with the convolution operation, it requires training a number of parameters proportional to the environment size.
%To reduce the number of trainable parameters and make the model more reliable, we replace such encoding with the explicit beacons' locations in the proposed memory-efficient PFRNN (mePFRNN).
To reduce the number of trainable parameters, we do not use the data about an environment as input to our model since such data, like beacons' and obstacles' positions, have to be implicitly extracted in the training stage.  
We expect such behavior of the considered mePFRNN since the environment is the external factor to the localization problem and stays the same over the particular trajectory.
% Such a network can be pre-trained once and further used for other environments.
% This allows to generalize the same network to be used it for various environments without re-training the model.
%In particular, instead of the encoding scheme~(\ref{eq::env_embed_pfrnn}), input for mePFRNN model is the raw measurement data $\rvy^{(k)}_i$ and embeddings of motion and measurement constructed as follows.
The motion and measurement vectors corresponding to every particle are embedded into a high dimensional space via linear layer and ReLU non-linearity.
Then, the obtained embeddings are concatenated and processed by a linear layer with LeakyReLU non-linearity.
The result of the latter operation is motion embedding $\rve^{(k)}_{\rvu}$ for every particle, which is additional input to the proposed mePFRNN cell.
The encoding procedure described above is summarized in scheme~(\ref{eq::mepfrnn_input}).
\begin{equation}
    \begin{array}{cc}
       \rvx_i^{(k)} \to \mathrm{Linear} \to \mathrm{ReLU} \; {}_{\searrow} &   \\
       & \mathrm{Concatenation} \to \mathrm{Linear} \to \mathrm{LeakyReLU} \to \rve^{(k)}_{\rvu} \\
        \rvy_i^{(k)} \to \mathrm{Linear} \to \mathrm{ReLU} \; {}^{\nearrow} & 
    \end{array}
    \label{eq::mepfrnn_input}
\end{equation}

Thus, mePFRNN is a voxel-independent model that can be easily used in very large environments without increasing the number of trainable parameters.
% Also, we can limit the number of used beacons' positions and focus only on the positions in the target neighborhood to improve the robustness of the model.
One more benefit of the proposed approach becomes crucial if the beacons in the environment are located not in the middle of the artificially generated voxels in the PFRNN model.
These voxels compose a grid for the considered environment to identify the beacons and obstacles with convolution encoding.
In this case, the convolution operation does not adequately encode the beacons' positions and makes further filtering more noisy.
The resulting cell is shown in Figure~\ref{fig:mepfrnn} graphically and in equations~(\ref{eq::mepfrnn}) analytically, where $MLP$ consists of two sequential linear layers and intermediate $\mathrm{LeakyReLU}$ nonlinearity. 
Note that, the Soft Resampling procedure is also used here similar to the PFRNN model described above.

\begin{minipage}{.4\textwidth}
\centering
\includegraphics[width=\textwidth]{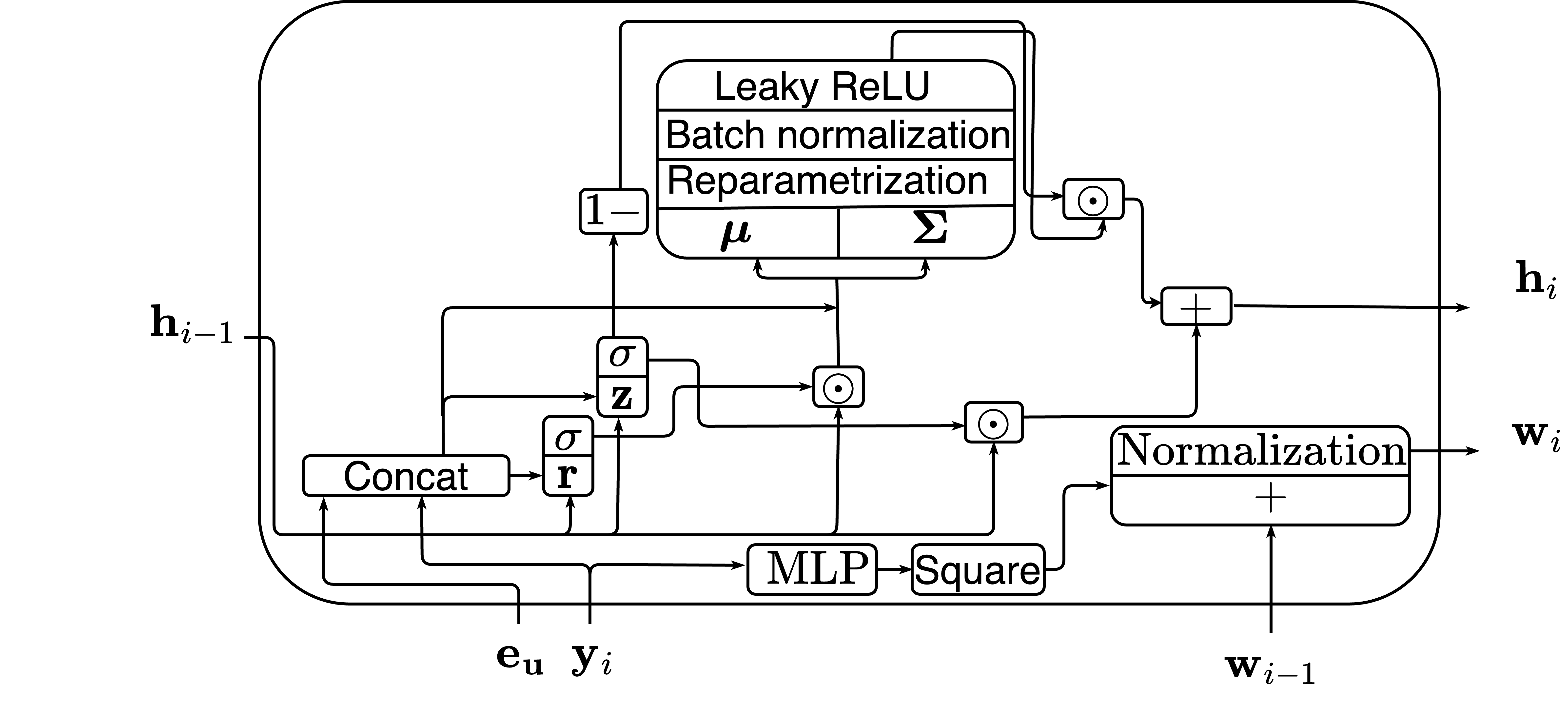}
\captionof{figure}{The proposed mePFRNN cell. Square block means elementwise square of the input. MLP consists of two sequential linear layers and LeakyReLU intermediate nonlinearity. Other notation is similar to the PFRNN.}
\label{fig:mepfrnn}
\end{minipage}
\begin{minipage}{.6\textwidth}
\begin{equation}
    \begin{split}
        \rve^{(k)}_i =& [\rve^{(k)}_{\rvu}, \rvy^{(k)}_i] \\
        \rvz^{(k)}_i = &\sigma(\mW_{z} [\rvh_{i-1}^{(k)}, \rve^{(k)}_i] + \rvb_z)\\
        \rvr^{(k)}_i = & \sigma(\mW_{r} [\rvh_{i-1}^{(k)}, \rve^{(k)}_i] + \rvb_r)\\
        \vmu^{(k)}_i =& \mW_{\vmu} [\rvr_i^{(k)} \odot \rvh_{i-1}^{(k)}, \rve^{(k)}_i] + \rvb_{\vmu} \\
        \mSigma^{(k)}_i =& \mW_{\mSigma} [\rvr_i^{(k)} \odot \rvh_{i-1}^{(k)}, \rve^{(k)}_i] + \rvb_{\mSigma} \\
        \rvepsilon \sim & \; \mathcal{N}(0, \mI)\\
        \rvd^{(k)}_i = & \mathrm{LeakyReLU} (\mathrm{BN}(\vmu^{(k)}_i + \mSigma^{(k)}_i \odot \rvepsilon))\\
        \rvh_i^{(k)} = & (1 - \rvz^{(k)}_i) \odot  \rvd^{(k)}_i + \rvz^{(k)}_i \odot \rvh_{i-1}^{(k)}  \\
        w_i = & MLP(\rvy_i^{(k)})^2 + w_{i-1} - \\
        & - LogSumExp(MLP(\rvy_i^{(k)})^2 + w_{i-1})
    \end{split}
    \label{eq::mepfrnn}
\end{equation}
\end{minipage}

% We illustrate the considered RNN cells in Figure~\ref{fig::rnn_comparison} for the reader's convenience. 
% The following main points of difference are implemented in the model- based models are the following:
% \begin{itemize}
%     \item The model is voxel independent since doesn't use world environment convolution and embedding. Instead, it uses the exact beacons' location as input data.
%     of the environment.
%     \item The target function treats angle error minimization differently: instead MSE of the angle in radians it uses sum of MSE of $\cos$ and $\sin$ of the angle.
% \end{itemize}
% The first difference leads to a fixed number of parameters independent of the complexity and size of the world. Moreover, practically it can easily be modified to very large worlds by using only part of the beacon's information in the vicinity of the interval of the interest (at least when the model converges to a state with some accuracy). 
% Also, it makes the model more reliable if the beacons are located not in the middle of the cells as the location is known exactly.
% The second difference is caused by the point that the angle $2\pi+\eps$ and $\eps$ physically are the same angles, while the MSE treats them as essentially different. The use of angle' $\sin$ and $\cos$ functions determines the direction explicitly.
% At the same time, these differences are important for the localization problem and are not essential for the stocks prediction or any other problems considered in \cite{ma2020particle}.

\paragraph{Alternative GRU-based models.}
In addition to the proposed mePFRNN model, we also propose two approaches to exploiting the classical GRU model (see Figure~\ref{fig:gru_rnn} and equations~(\ref{eq::gru})) in the object localization problem.
Namely, the EnsembleGRU model consists of many small GRU cells whose predictions are averaged to  estimate the target object state.
The number of models in the ensemble and the number of trained parameters in every model are selected such that the total number of the trained parameters is approximately equal to \# parameters in PFRNN times \# particles.
The complementary approach is just to use the single GRU cell, where the number of trained parameters is equal to \# particles times \# parameters in PFRNN.
Both approaches are complementary to the PFRNN and mePFRNN models since they do not exploit particles.
Also, note that the input to the GRU cell in EnsembleGRU and HeavyGRU models is the same as the input to the PFRNN cell.

\begin{minipage}{.5\textwidth}
\centering
\includegraphics[width=0.9\textwidth]{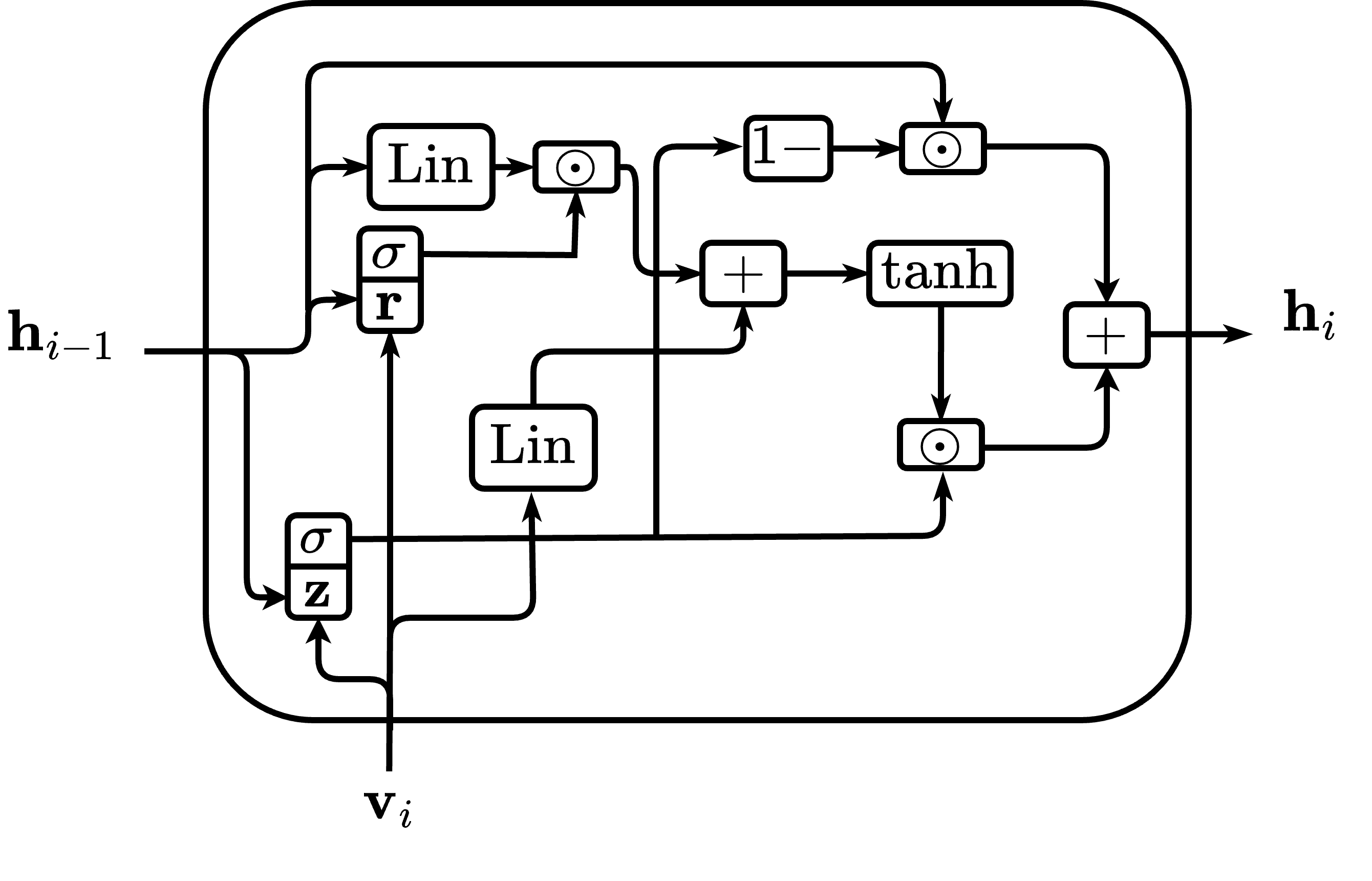}
\captionof{figure}{Standard GRU cell, where $\rvz$ and $\rvr$ denote the linear layers to compute $\rvz_i$ and $\rvr_i$, $\mathrm{Lin}$ denotes linear layers to compute $\hat{\rvh}_i$, see~(\ref{eq::gru}).}
\label{fig:gru_rnn}
\end{minipage}
\begin{minipage}[b]{.4\textwidth}
\begin{equation}
  \begin{aligned}
  \rvz_i = &\sigma(\mW_{z}[\rvv_i, \rvh_{i-1}] + \rvb_z) \\
\rvr_i = &\sigma(\mW_{r} [\rvv_i, \rvh_{i-1}] + \rvb_r) \\
\hat{\rvh}_i = &\tanh(\mW_{h} \rvv_i + \rvb_h + \\
& \rvr_i \odot (\mU_{h} \rvh_{i-1} + \rvb_u)) \\
\rvh_i =&   (1-\rvz_i) \odot \rvh_{i-1} + \rvz_i \odot  \hat{\rvh}_i
  \end{aligned} 
  \label{eq::gru}
  \end{equation}
\end{minipage}
\section{Computational experiment}
\label{sec::experiment}

In this section, we demonstrate the performance of our model and compare it with alternative neural networks and non-parametric models.
For training the compared neural networks we use RMSProp optimizer~\cite{tieleman2012rmsprop} since it shows more stable convergence compared to Adam~\cite{kingma2014adam} and SGD with momentum~\cite{goodfellow2016deep}, learning rate equal to $5\cdot 10^{-4}$ and every batch consists of 150 trajectories.
% The loss function used for PFRNN and GRU models consists of the coordinate MSE and angle MSE: $(x-x_0)^2+(y-y_0)^2+\beta(\phi-\phi_0)^2$ with the weight $\beta=0.1$. 
% In the MB-PFRNN the angles loss function was chosen as a combination of $(x-x_0)^2+(y-y_0)^2+\beta(\sin\phi-\sin\phi_0)^2+\beta(\cos\phi-\cos\phi_0)^2$ with the same $\beta$ coefficient. 
% This modification is done to treat correctly angles outside $2\pi$ values.
The maximum number of epochs is 5000 for the considered environments.
During the training stage, a validation set of trajectories is used to identify the overfitting.
Therefore, different environments require a different number of epochs before overfitting occurs.
In particular, overfitting does not occur after 5000 epochs in the world $10\times 10$.
At the same time, overfitting is observed after 600 and 200 epochs in the World $18\times 18$ and WORLD $27 \times 27$, respectively.
% Such a number of epochs is enough to reach minimal values for the evaluation set at the world $10\times 10$; for the World $18\times 18$ and WORLD $27 \times 27$ typically 600 and 200 epochs respectively lead to overfitting.
%Such difference is due to the overfitting phenomenon observed in the case of %ultra-symmetric environments like WORLD $27 \times 27$. 

\begin{figure}[!h]
    \centering
    \begin{subfigure}{0.3\textwidth}
    \includegraphics[width=\textwidth]{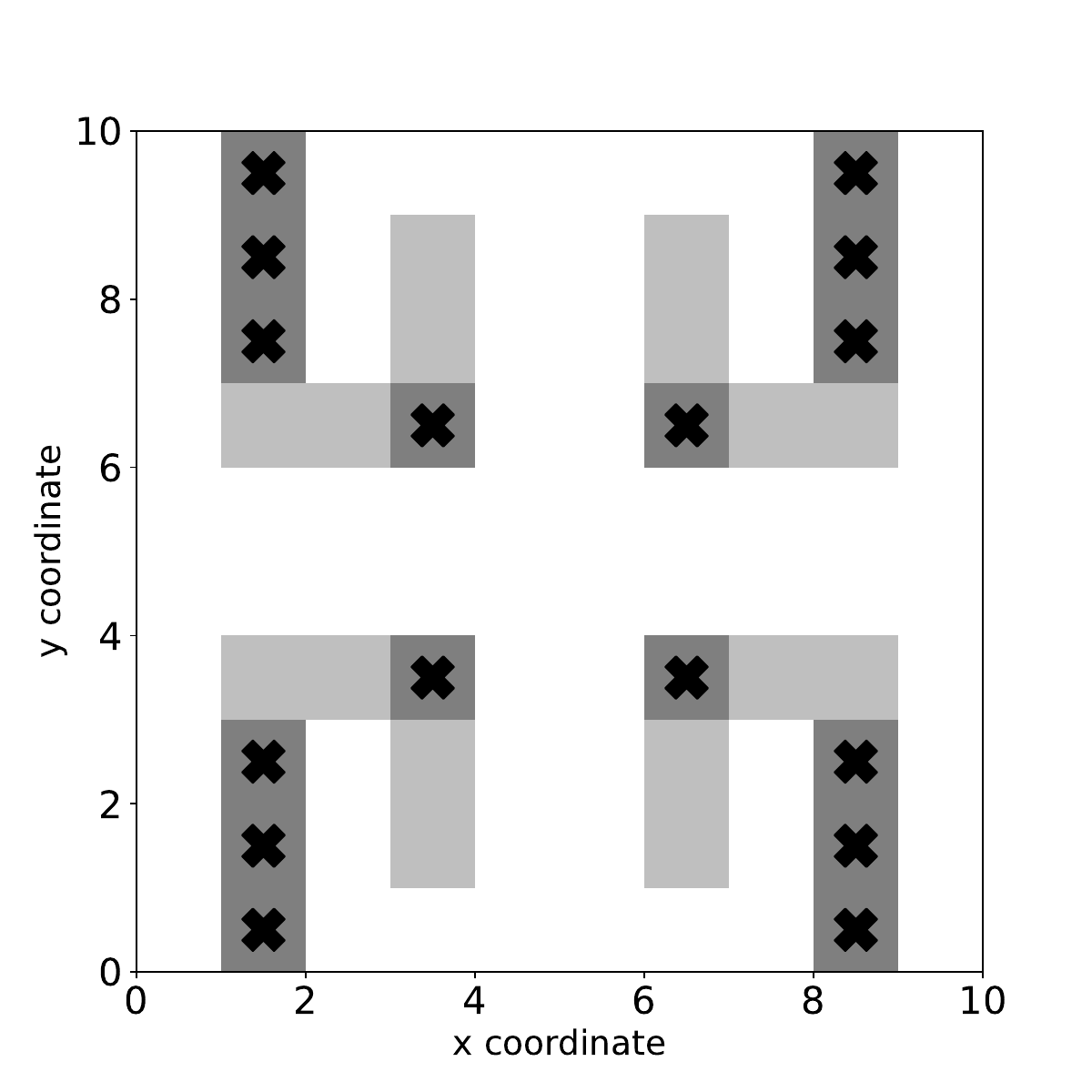}
    \subcaption{world $10 \times 10$}
    \end{subfigure}
    ~
    \begin{subfigure}{0.3\textwidth}
    \includegraphics[width=\textwidth]{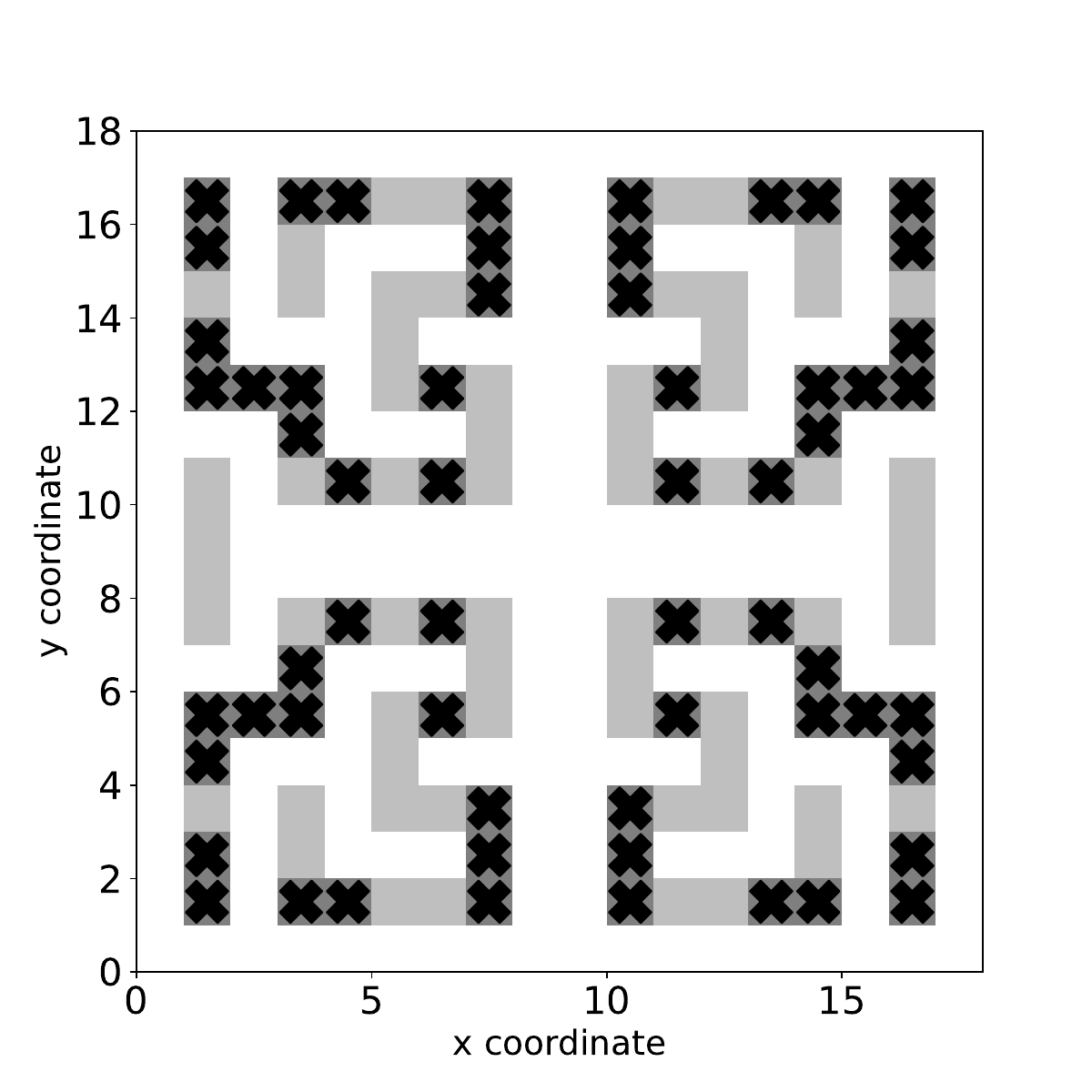}
    \subcaption{World $18 \times 18$}
    \end{subfigure}
    ~
    \begin{subfigure}{0.3\textwidth}
    \includegraphics[width=\textwidth]{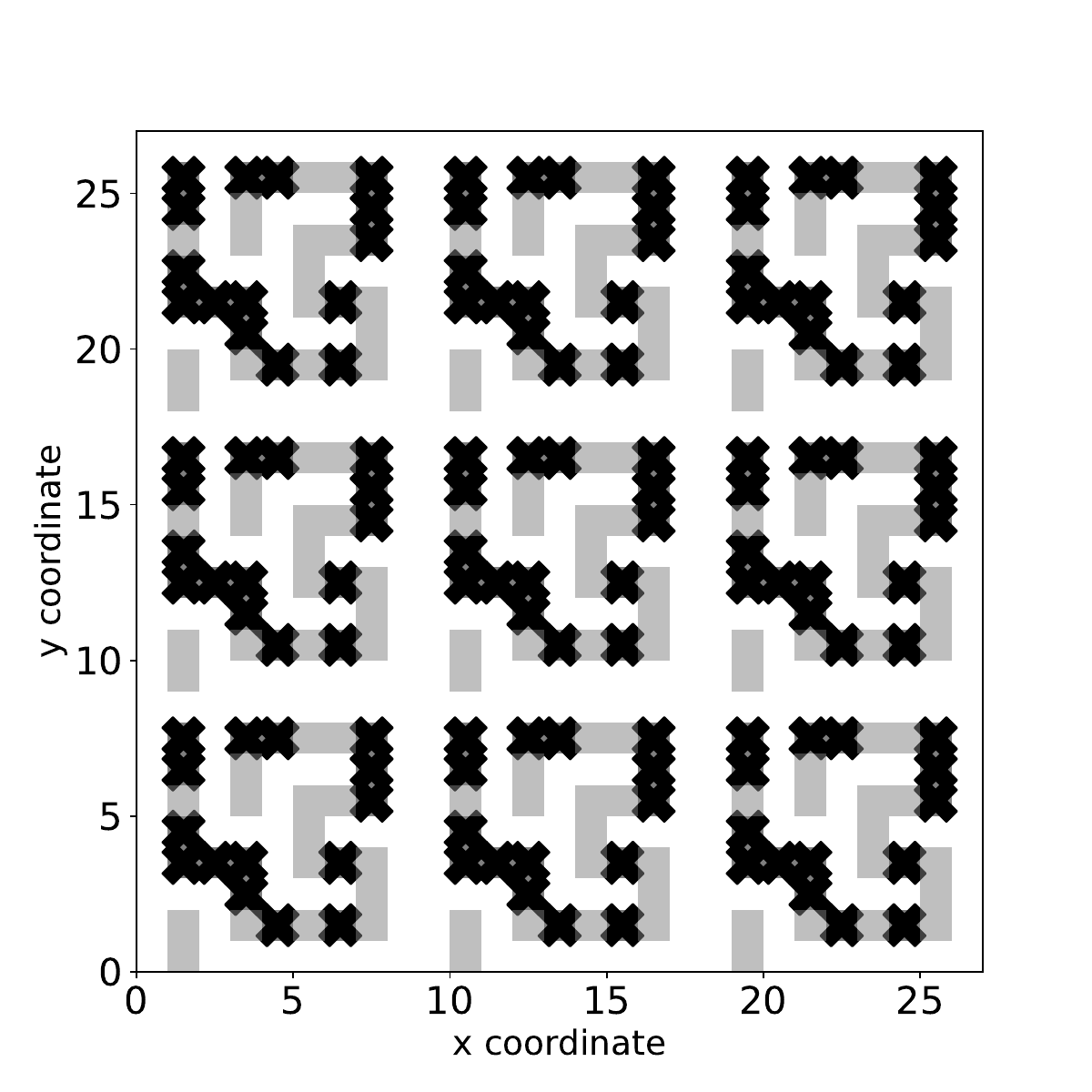}
    \subcaption{WORLD $27 \times 27$}
    \end{subfigure}
    \\
    \begin{subfigure}[t]{0.4\textwidth}
    \includegraphics[width=\textwidth]{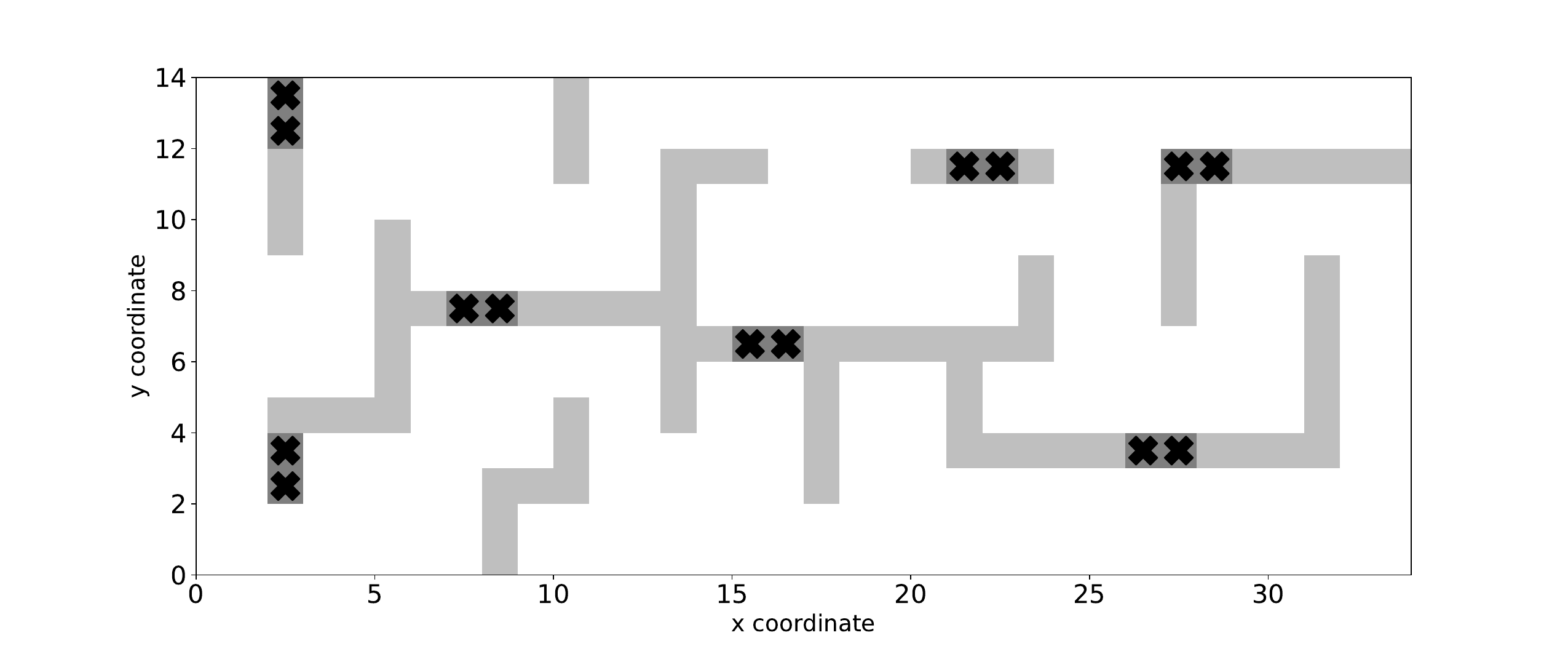}
    \subcaption{\emph{Labyrinth}}
    \end{subfigure}
    \caption{Visualization of test environments. Black crosses denote beacons, and grey blocks denote obstacles. The upper row represents the very symmetric environments that are especially challenging for solving the localization problem. The \emph{Labyrinth} environment is not symmetric and is similar to the environment, which was used for the evaluation of filtering methods in~\cite{zhu2020towards}.}
    \label{fig::test_environments}
\end{figure}

\paragraph{Trajectories generation procedure.}
To evaluate the considered methods and demonstrate the performance of the proposed mePFRNN, we consider four environments, see Figure~\ref{fig::test_environments}.
Environments world $10 \times 10$, World $18 \times 18$,  and WORLD $27 \times 27$ are symmetric and therefore challenging for object localization since symmetric parts can be confused by a filtering method.
Environment \emph{Labyrinth} is not symmetric and medium challenging for filtering methods.
Thus, the considered filtering methods are compared comprehensively due to the diversity in the testing environments.

To train the parametric models we need to generate a set of trajectories $\{\rvx_i^*\}_{i=1}^N$.
Since our tests assume that the object's initial state is unknown, we set the initial state $\rvx_0^*$ randomly for all generated trajectories.
Initial states do not intersect with obstacles.
Then, every next iteration updates the object state according to the motion equation, where external velocity $u \in [0, 0.2]$ is known and the direction is preserved from the previous step within the noise.
In the case of a collision with an obstacle, the object's direction is changed randomly such that the next state does not indicate the collision.
To simulate engine noise, the velocity~$u$ is perturbed by $\eta_r \sim \mathcal{U}[-0.02, 0.02]$.
To simulate uncertainty in the object control system, the direction $\phi$ is also perturbed by $\eta_{\phi} \sim 2\pi \alpha$, where $\alpha \sim \mathcal{U}[-0.01, 0.01]$. 
The measurements $\rvy_i$ are the distances to the five nearest beacons, which are also noisy with the noise distributed as $\zeta \sim \mathcal{U}[-0.1, 0.1]$.
In the considered environments, we set the number of time steps in every trajectory $N=100$. 

To train the considered parametric models, we generate 8000 trajectories, 1000 trajectories for validation, and an additional 10000 trajectories for the testing stage.
During the training process, the MSE loss is computed for the validation trajectories and if the obtained value is smaller than the current best one, then the best model is updated.
This scheme helps to store the best model during the training and avoid overfitting.

\paragraph{The list of compared models.}
We compare the proposed mePFRNN model with the following competitors combined in two groups.
The first group consists of alternative recurrent neural networks that can solve the object localization problem, in particular the baseline PFRNN model from~\cite{ma2020particle}, HeavyGRU, and EnsembleGRU models.
% The latter two models are constructed from the standard GRU cell, but they differ in the number of trained parameters since there are different ways to incorporate particles from PFRNN into the standard GRU cell.
% The HeavyGRU model is the standard GRU cell, where the number of trained parameters is equal to \# particles times \# parameters in PFRNN.
% The EnsembleGRU model works as an ensemble of $p$ GRU models, where $p$ is the number of particles in PFRNN and every GRU model from the ensemble has the same number of parameters as every particle in PFRNN. 
Following the study~\cite{ma2020particle} we use the $wMSE$ loss function~\eqref{eq::wmse_def} to train alternative neural network models and use $L$ loss function~\eqref{eq::L_def} to train the proposed mePFRNN model.
Such a choice of training setup highlights the benefit of the proposed loss function $L$.
In both settings, we use $\beta = 0.1$.

The second group consists of the particle filter (PF) and the multiparticle Kalman filter (MKF).
We include these methods in the experiments to compare the performance of the parametric and non-parametric models.
The performance is measured in terms of MSE$_c$, FSE, number of trained parameters, training time, and inference time.
Note that, non-parametric models do not require training, therefore they are more lightweight.
However, to get high accuracy a lot of particles are needed which leads to long runtime.
Thus, for adequate comparison with neural methods, the classical filters were used with fewer particles to show a similar runtime as neural network-based models in the inference mode.
In addition, we use stochastic resampling in the non-parametric models and Soft Resampling in the parametric ones.
However, the Soft Resampling procedure for the non-parametric models does not significantly change the final performance.
The comparison of the aforementioned models is presented in the next paragraph.

\paragraph{Discussion of the results.}

In experiment evaluation, we compare non-parametric and parametric models with the four test environments described above.
The obtained results are summarised in Table~\ref{tab::perf_comp_nn_models}.
Also, we track the number of trained parameters, the amount of memory that is necessary to store them, and the runtime to update the object state in one step.
From this table follows that the proposed mePFRNN model gives the best or the second-best MSE$_c$ score for the considered environments.
At the same time, the FSE score is typically smaller for HeavyGRU or EnsembleGRU in the considered environments.
One more important factor is the number of trainable parameters.
The smaller the number of parameters, the easier embedding the model in hardware.
The mePFRNN model requires fewer trainable parameters compared with other parametric models, i.e. PFRNN, HeavyGRU, and EnsembleGRU.
The last but not least feature of the considered models is the inference time, i.e. the runtime to update the object state from the $i$-th to the $(i+1)$-th time step.
mePFRNN is slightly faster than PFRNN, and HeavyGRU appears the fastest model in the inference stage. 
Thus, we can conclude that the proposed mePFRNN model provides a reasonable trade-off between MSE$_c$ score, number of trainable parameters, and inference time among the considered parametric and non-parametric models tested in the selected benchmark environments.

\begin{table}[!h]
    \centering
\caption{Performance comparison of the filtering methods. Mean values averaged over 10000 runs and standard deviations in braces. 
The number of particles for PF and MKF is selected such that their filtering time is close to the inference time in  neural network models. 
Dashes indicate non-parametric models, which do not have any trainable parameters and therefore do not consume memory. 
Note that we report MSE$_c$ and FSE values corresponding to object position only and ignore the angle component of the state vector.}
\begin{adjustbox}{width=\columnwidth,center}
    \begin{tabular}{cccccccc}
    \toprule
        Environment & Model & MSE$_c$ & FSE & \# parameters & \# particles & Memory, Mb & Inference time, ms.  \\ 
        \midrule
        \multirow{6}{*}{world $10 \times 10$} & %mePFRNN (our) & $\mathbf{0.77 \; (3.97)}$ & 0.13	(0.09) & 28802 & 30 & 0.46 & 1.4\\
        mePFRNN (our) & $\mathbf{0.87 \; (4.68)}$ & 0.13	(0.16) & 28802 & 30 & 0.46 & 1.0\\
        & PFRNN & 1.30	(5.70) & 0.05 (0.18) & 99472 & 30 & 1.5 & 1.1 \\
        & HeavyGRU & 1.01	(5.31) & 0.07 (0.13) & 2453239 & 1 & 37 & 0.4 \\
        & EnsembleGRU & 1.24 (4.49) & $\mathbf{0.04	(0.13)}$ & 95283 & 30 & 45  & 4.4\\ 
        & PF & 13.70	(25.47) & 1.73 (2.81) & $-$ & 200 & $-$ & $2.0$ \\
        & MKF & 10.77 	(23.87) & 1.37 (2.60) & $-$ & 50 & $-$ & $4.7$ \\
        \midrule
        \multirow{6}{*}{World $18 \times 18$} &
        %mePFRNN (our) & 7.08 (25.69) & 0.51 (0.63) & 28802 & 30 & 0.46 & 1.5\\
        mePFRNN (our) & 6.89 (23.86) & 0.51 (0.63) & 28802 & 30 & 0.46 & 1.0\\
        & PFRNN & 10.74	(29.57) & 0.30	(0.71) & 214160 & 30 & 1.6 & 1.1\\
        & HeavyGRU & $\mathbf{ 6.83	\; (26.13)}$ & $\mathbf{0.22	(0.49)}$ & 2682615 & 1 & 41 & 0.4 \\
        & EnsemleGRU & 9.79 (22.04) & 0.24 (0.54) & 209971 & 30 & 99 & 4.3 \\ 
        & PF & 74.17	(91.80) & 5.73	(5.37) & $-$ & 200 & $-$ & $2.4$ \\
        & MKF & 96.13	(114.14) & 7.08	(6.57) & $-$ & 50 & $-$ & $4.6$ \\
        \midrule
        \multirow{6}{*}{WORLD $27 \times 27$} &
        %mePFRNN (our) & $\mathbf{59.65 \;	(55.59)}$ & $\mathbf{5.52 \; (3.76)}$ & 28802 & 30& 0.46 & 1.5\\
        mePFRNN (our) & $\mathbf{61.94 \;	(55.47)}$ & $\mathbf{5.69 \; (3.79)}$ & 28802 & 30& 0.46 & 1.0\\
        & PFRNN & 68.28	(64.16) &  5.86	(3.96) & 465392 & 30 & 7.1 & 1.1\\
        & HeavyGRU & 73.36	(67.87) & 6.22	(3.99) & 3169367 & 1 & 48 & 0.4\\
        & EnsembleGRU & 67.41	(59.13) & 5.86 (3.67) & 461203 & 30 & 220 & 4.3 \\ 
        & PF & 181.75	(171.81) & 11.09 (6.76) & $-$ & 200 & $-$ & $2.8$ \\
        & MKF & 200.36 (201.07) & 12.02 (7.60) & $-$ & 50 & $-$ & $6.8$ \\
        \midrule
        \multirow{6}{*}{$Labyrinth$} & %mePFRNN (our) & $\mathbf{1.43 \;	(13.27)}$ & 0.30 (0.24) & 28802 & 30 & 0.46 & 1.4\\
        mePFRNN (our) & $\mathbf{1.43 \;	(13.27)}$ & 0.30 (0.24) & 28802 & 30 & 0.46 & 1.0\\
        & PFRNN & 6.26	(29.28) &  0.18	(0.11) & 307696 & 30 & 2.5 & 1.1\\
        & HeavyGRU & 1.78	(13.80) & $\mathbf{0.12\;	(0.11)}$ & 2838263 & 1 & 45 & 0.4\\
        & EnsembleGRU & 5.57 (5.66) & 0.12 (0.08) & 303507 & 30 & 135 & 4.4\\ 
        & PF & 87.23 (163.00) & 4.74 (6.92) & $-$ & 200 & $-$ & $1.8$\\
        & MKF & 77.90 (169.20) & 4.19 (7.40) & $-$ & 50 & $-$ & $4.4$\\
        \bottomrule
    \end{tabular}
    \end{adjustbox}
    \label{tab::perf_comp_nn_models}
\end{table}

The number of particles chosen in Table~\ref{tab::perf_class_pf_mkf} is such that the inference runtime is close to the inference runtime of the considered neural networks. 
% When employing an optimal and substantial number of particles, the results significantly improve \ref{tab::perf_class_pf_mkf}. 
Since in Table~\ref{tab::perf_comp_nn_models} we fix the particular number of particles in non-parametric models, we present the MSE$_c$ and FSE losses for the larger number of particles in Table~\ref{tab::perf_class_pf_mkf}.
It shows that if the number of particles is sufficiently large, both MSE$_c$ and FSE values are smaller than the corresponding values for parametric models.
However, such an accurate estimation of states requires a much slower inference runtime compared to the considered parametric models.
Thus, the neural network-based filters are of significant interest since they can show better accuracy compared to non-parametric models and provide faster updates of the object's state.

\begin{table}[!h]
    \centering
    \caption{Dependence of the PF and MKF performance and inference time to update the state vector on the number of particles. The more particles are used in these filters, the more accurate trajectories are recovered and the slower filtering is. 
    Here we focus only on the loss functions that evaluate the accuracy of object coordinates filtering.}
    \begin{tabular}{ccccccc}
    \toprule
    Environment & Filter & \# particles & MSE$_c$ & FSE & Inference time, ms. \\
    \midrule
    \multirow{4}{*}{world $10\times 10$} & PF  & 200 & 13.70	(25.47) & 1.73 (2.81) & $0.8$ \\
        & MKF & 50 & 10.77 	(23.87) & 1.37 (2.60) & $4.7$ \\
        & PF  & 10000 & 0.58 (3.67) & 0.20 (0.09) & 45 \\
        & MKF & 10000 & 0.82 (3.75) & 0.20 (0.03) & 58 \\
    \midrule
    \multirow{4}{*}{World $18\times 18$} & PF & 200 & 74.17	(91.80) & 5.73	(5.37) & 2.4 \\
        & MKF & 50 & 96.13	(114.14) & 7.08	(6.57) & 4.6\\
        & PF & 10000 & 3.04 (18.24) & 0.22	(0.51) & 48 \\
        & MKF & 10000 & 3.25 (13.89) & 0.21	(0.11) & 66 \\
    \midrule
    \multirow{4}{*}{WORLD $27 \times 27$} & PF & 200 & 181.75	(171.81) & 11.09 (6.76) & 2.8 \\
        & MKF & 50 & 200.36 (201.07) & 12.02 (7.60) & 6.8 \\
        & PF & 10000 & 74.86 (79.48) & 5.94	(5.06) & 37 \\
        & MKF & 10000 & 55.83 (52.68) & 5.16 (3.93) & 50 \\
    \midrule
    \multirow{5}{*}{$Labyrinth$} & PF & 200 & 87.23 (163.00) & 4.74 (6.92) & 1.8 \\
        & MKF & 50 & 77.90 (169.20) & 4.19 (7.40)  & 4.4 \\
        & PF & 10000 & 1.53	(15.06) & 0.50	(0.02) & 40 \\
        & MKF & 10000 & 1.50 (14.81) & 0.50	(0.03) & 90 \\
        \bottomrule
    \end{tabular}
    \label{tab::perf_class_pf_mkf}
\end{table}

\section{Conclusion}

We present the novel recurrent neural network architecture mePFRNN to solve the object localization problem.
It combines the standard GRU RNN, particle filter, and explicit measurements of distances from the object to the beacons.
The latter feature makes the proposed model memory-efficient since the number of trainable parameters does not depend on the environment size.
% Also, the proposed mePFRNN is specifically important for the localization problem and is not essential for stock prediction or any other applications dealing with sequential data addressed by the PFRNN model.
We compare the proposed mePFRNN model with the general-purpose PFRNN model and two modifications of standard GRU RNN.
The test environments consist of symmetric environments of different sizes and the non-symmetric \emph{Labyrinth} environment.
Such diversity of the test environments leads to the comprehensive comparison of the considered parametric models to solve the object localization problem.
% Although mePFRNN appears slightly slower in inference than the baseline PFRNN, it filters the object's coordinates much more precisely in the considered symmetric environments.
The mePFRNN model is simultaneously   slightly faster in inference than the baseline PFRNN and filters the object's coordinates more precisely in the considered symmetric environments along the trajectory.
Moreover, mePFRNN does not exploit explicit data about the environment or the corresponding embeddings.
At the same time, the proposed mePFRNN model outperforms competitors in MSE values for the most of considered test environments.

\bibliographystyle{unsrt}

\bibliography{main.bib}

\end{document}